\documentclass[letterpaper]{article}

\usepackage{natbib,alifeconf}  

\usepackage{subcaption}
\usepackage{enumitem}  
\setlist{noitemsep}  
\usepackage{gensymb}  
\usepackage{hyperref}
\usepackage{xcolor}
\definecolor{dark-green}{rgb}{0,0.5,0}

\title{Artificial Plants \textendash{} Vascular Morphogenesis Controller-guided \\growth of braided structures}

\author{Daniel N. Hofstadler$^{1}$, Joshua C. Varughese$^{1, 2}$, Stig A. Nielsen$^3$, David A. Leon$^{4, 5}$, \\{\Large Phil Ayres$^4$, Payam Zahadat$^{1}$ \and Thomas Schmickl$^1$} \\
\mbox{}\\
$^1$University of Graz, Austria \\
$^2$Graz University of Technology, Austria \\
$^3$IT University of Copenhagen, Denmark \\
$^4$The Royal Danish Academy of Fine Arts, Denmark \\
$^5$LEAP, Spain\\
daniel.hofstadler@uni-graz.at} 

\begin{document}
\maketitle


\begin{abstract}
Natural plants are exemplars of adaptation through self-organisation and collective decision making.
As such, they provide a rich source of inspiration for adaptive mechanisms in artificial systems.
Plant growth \textendash{} a structure development mechanism of continuous material accumulation that expresses encoded morphological features through environmental interactions \textendash{} has been extensively explored in-silico.
However, ex-silico scalable morphological adaptation through material accumulation remains an open challenge.
In this paper, we present a novel type of biologically inspired modularity, and an approach to artificial growth that combines the benefits of material continuity through braiding with a distributed and decentralised plant-inspired Vascular Morphogenesis Controller (VMC).
The controller runs on nodes that are capable of sensing and communicating with their neighbours.
The nodes are embedded within the braided structure, which can be morphologically adapted based on collective decision making between nodes.
Human agents realise the material adaptation by physically adding to the braided structure according to the suggestion of the embedded controller.
This work offers a novel, tangible and accessible approach to embedding mechanisms of artificial growth and morphological adaptation within physically embodied systems, offering radically new functionalities, innovation potentials and approaches to continuous autonomous or steered design that could find application within fields contributing to the built environment, such as Architecture.
\end{abstract}

\section{Introduction}

Adaptive behaviours of natural plants have been subject to intensive study through in-silico model building,
garnering valuable insights as {\it models} of underlying biological processes.
Many of these models find further utility as {\it models for} driving and controlling artificial systems, and often involve transfers across disciplinary fields.
As a particular case, computationally focused Architectural design, has consistently sought to transfer plant inspired models, growth models in particular, to act as generative design engines. For example, L-systems~\citep{lindenmayer1968mathematical} have been used as space planning and form-giving generators~\citep{coates2001current, serrato2005lindenmayer, fernando2015recapitulation}, whilst leaf venation models~\citep{runions2005modeling} have been employed as growth models informing the morphology of experimental architectural installations~\citep{tamke2013rise}.
It is of interest to note, however, that within the architectural design process, the dynamics of such models are arrested once design goals have been achieved \textendash{} morphological adaptation occurs in-silico, but these mechanisms do not transfer into physical embodiment and context, where real adaptation could occur.
One contributing factor to this is that architectural practice is focused on the production of `end-points', or `completed' buildings, structures and spaces~\citep{burry2013designing}.
Another factor is that ex-silico morphological adaptation presents a significant challenge to conventional construction technologies that do not easily facilitate continuous change and alteration.
In this paper, we present a novel approach to embodied artificial growth that combines the benefits of material continuity using braiding technique with a plant-inspired distributed and decentralised Vascular Morphogenesis Controller (VMC).
Although our demonstration does not approach architectural scale, the underlying principles of construction and control are scalable. 

The work presented here contributes to the project {\it flora robotica}~\citep{hamann2015florarobotica, hamann2017florarobotica}
which explores the symbiotic coupling of plants and robots to grow continuously adaptive bio-hybrid structures towards architectural objectives.
The technique of braiding is used to construct scaffolds that are populated with robotic nodes for steering plant growth to desired targets, and for plants to grow upon.
Braided structures are highly suited for this purpose; they consist of fibres in a reciprocally interlaced configuration giving them the flexibility to change topology, i.e., bifurcate, merge and extend in different directions.
Additionally, these structures are well suited for incorporation of wires and electronics necessary for feedback and control devices.
Braids also possess useful mechanical properties; they can expand/shrink and change shape under pressures while still keeping their integrity; they can be stiffened/weakened by adding/removing filaments on site.

Coupling these properties with the embedded VMC controller running on distributed nodes creates an embodied system that can demonstrate morphological adaptation with attributes of continuous growth. The nodes collectively decide from which locations the structure should grow further. The decision is signaled to a human agent who realises the actual growth/modifications by directly braiding the structure. The pre-determined parameters of the controller (in analogy to the genome of a plant), the status of the structure, and the environmental conditions direct the decision for the growth. In the following, we introduce core concepts underlying our approach, the design of the VMC, its integration with the embodied braid system, our experimental setup, results and conclusions.

\begin{figure}
\centering
\includegraphics[width=0.5\textwidth]{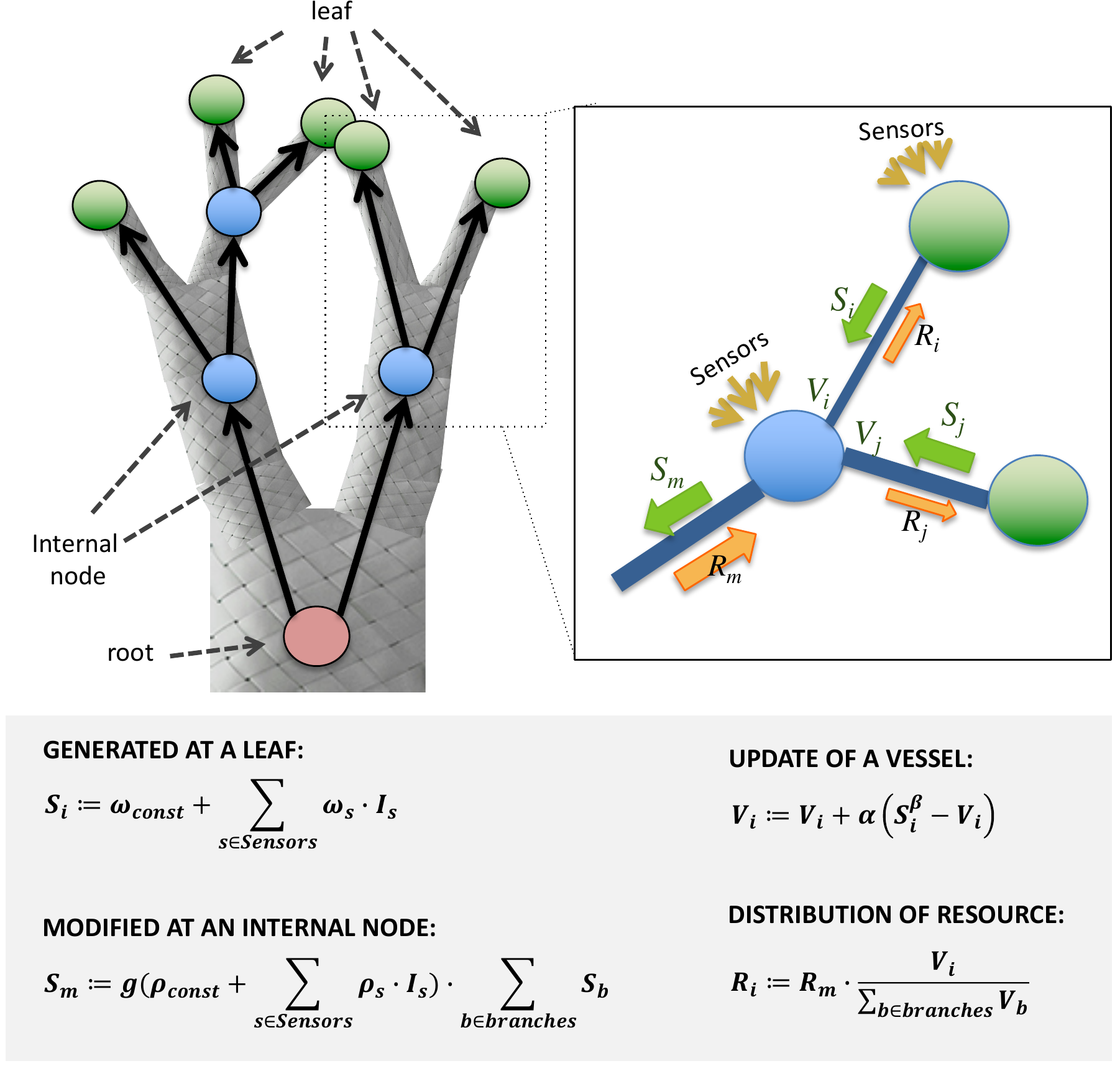}
\caption{Illustration and equations of the VMC algorithm. It is presented in detail in \cite{zahadat2017a}.
The parameterization used in this work is given in table~\ref{tab:vmc-constants}.}
\label{fig:VMC}
\end{figure}

\subsection{Biologically inspired modularity}

Plants are genetically evolved to adaptively grow by means of a modular organization~\citep{barthelemy2007architecture}.
The basic structural module is the \emph{phytomer},
consisting of the leaf-attachment-site (where branching can occur), the leaf itself, and the part of the stem leading to the next leaf.
However, this modularity is driven from processes of incremental material accretion that results in a materially integrated structural whole.
Our approach draws inspiration from this nuanced biological modularity by defining modules of localised function, but where material has structural continuity that contributes to the whole through braiding.
Returning to natural plants, phytomer by phytomer, a plant shoot optimizes its interactions with the environment by communicating via the vascular system which operates as a competitive distributor of resources~\citep{lucas2013plant, Notaguchi2015}.

\subsection{Competition for a common resource via vascular system dynamics}

Water, nutrients and hormones are distributed from scattered sources to sinks across the plant.
Water is typically pulled from roots toward leaves, where \textendash{} due to the requirements of photosynthesis \textendash{} gas-exchange drives evaporation.
The strongest sinks for the organic material provided by photosynthetically active leaves are growing tips.  


These in turn are the primary source of the main plant patterning hormone \emph{auxin}~\citep{Leyser2011}, that is transported rootward.
The \emph{canalization} hypothesis~\citep{sachs1981canalization, bennett2014flux} states that more `successful' branches produce more auxin, a greater flow of auxin causes increased production of vessels and indeed formation of better connections to the nearest auxin-`highways'.
Well-growing branches thus secure ever greater access to the pool of shared resources, which is necessary to sustain growth.


On the flipside, this mechanism turns all (potential) growth points \textendash{} on the same genetic individual \textendash{} into competitors for the common resource.
Exploring different growth options and developmentally promoting the most promising ones is highly adaptive in dynamic, unpredictable environments.
Such structures optimally exploit their local situation while requiring minimal genetic information, because of modularity~\citep{shinohara2013branching}).

\subsection{Vascular Morphogenesis Controller (VMC) algorithm}

In analogy to plant phytomers, a VMC-\emph{module} has an orientation and consists of a number of \emph{controller-nodes} (hereafter called \emph{nodes}): one \emph{root}-node and, representing the growth options of the module, some \emph{leaf}-nodes.
Each of the leaf-nodes of a VMC-module can merge with the root-node of another VMC-module (a \emph{child}-module) to become an \emph{internal}~node of the thus growing VMC-system or -graph (see Fig.~\ref{fig:VMC} for a sketch).

Each controller-node contains a set of state variables, whose dynamics are maintained locally by a set of rules, the input values coming from neighboring nodes and local sensors, as well as a set of constant parameters, the \emph{genome} of the VMC-system.
The genome is identical for all the nodes of the network and is subject to optimization (e.g., evolutionary algorithms).

The state variables in a controller-node are:
\begin{itemize}
\item Resource $R$: indicates a node's capability of new growth (i.e., it models supply in water, mineral nutrients, sugars, etc.)
\item Success $S$: indicator of how well a branch is doing, analogous to the plant hormone auxin.
\item Vessels $V_1$, $V_2$, ..., $V_n$: combined cross-sections or conductivity of vessels in the edges toward the $n$ children.
\end{itemize}

In analogy to the plant hormone auxin,
$S$ is compiled at the growing apices (the leaves of the graph) depending on local sensor values.
Up to the root(s) each node reports its value of $S$ to its parent-node, where all incoming values of $S$ are summed up and passed on to the next parent.
Along its way, it is altered at every step by local sensors and the genomic constants.
Crucially, incoming $S$ determines the future distribution of resources (variable $R$) by positively influencing the quantity of vessels
between two nodes.
Resource $R$ is generated at the root(s) and distributed downstream toward the leaves according to the relative thickness of the vessels $V_i$ at each junction.
New growth (i.e., the addition of a VMC-module) preferably happens at the leaves with the highest $R$ (see Fig.~\ref{fig:VMC} for a summary of the algorithm and the relevant equations).

Here we introduce flexible hardware and run the VMC algorithm as a showcase.

\section{Hardware Setup}

One Raspberry Pi\footnote{https://www.raspberrypi.org/products/raspberry-pi-3-model-b} (RPi) controls each single ``RasPiNet'' (or RPN) module.
For simplicity, each module only has two leaf-nodes, to which other modules can attach as children.
The structural basis is a Y-shaped braided module
that can be deformed in various ways.
The electronic parts are attached to the surface.
An RPi with an add-on board (the \emph{root-node}) near the fork of the braid contains plugs for parent-modules.
At the top of each of the arms of the Y, there is a sensor-board (the \emph{leaf-node})
that also serves as the electronic interface to a child-module's RPi.


\subsection{Electronic parts}

We designed two types of PCBs to allow compact and flexible deployment of the electronic components to the braided structures.
We call them \emph{Root Node} (RN) and \emph{Leaf Node} (LN), and describe them in the following.


\begin{figure}[htb]
\centering
\includegraphics[width=0.48\textwidth]{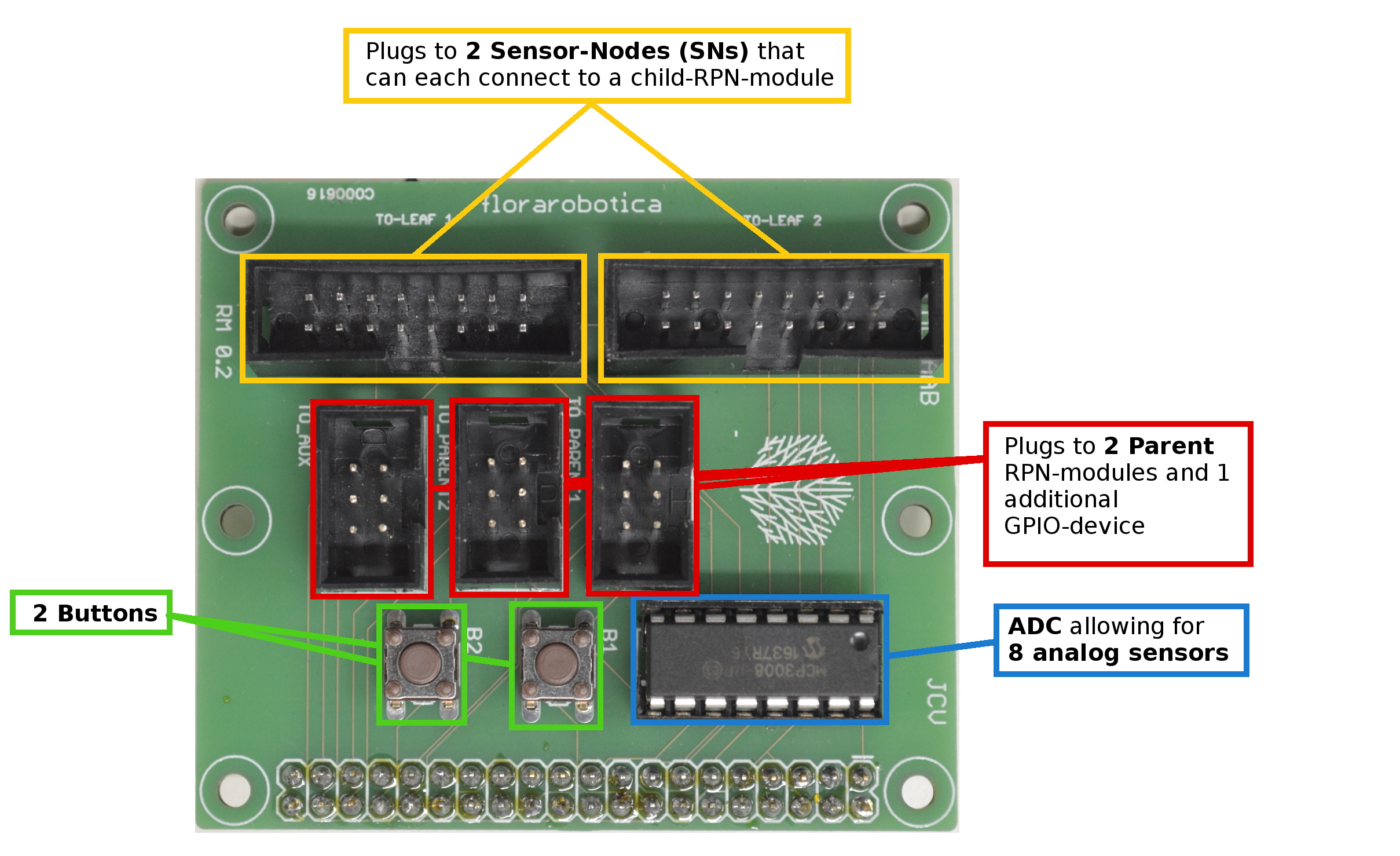}
\caption{Assembled root node (RPi not shown).}
\label{fig:rpn-root-board}
\end{figure}

\begin{figure}[htb]
\centering
\includegraphics[width=0.48\textwidth]{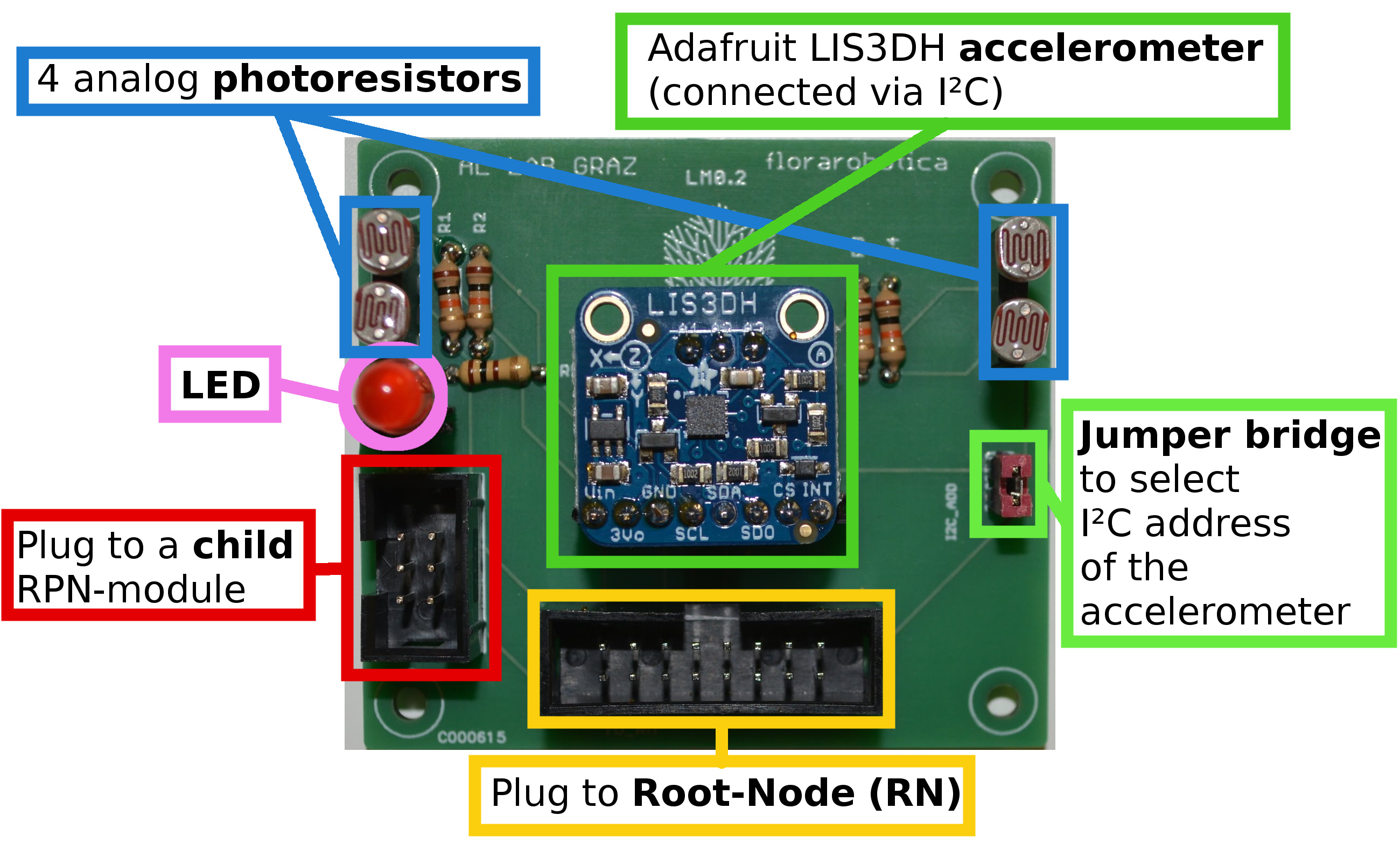}
\caption{Assembled leaf node. When attached to the braid, the four photo resistors are plugged via cables and distributed around the perimeter of the tube.}
\label{fig:rpn-leaf-board}
\end{figure}

\begin{figure}[htb]
\centering
\includegraphics[width=0.48\textwidth]{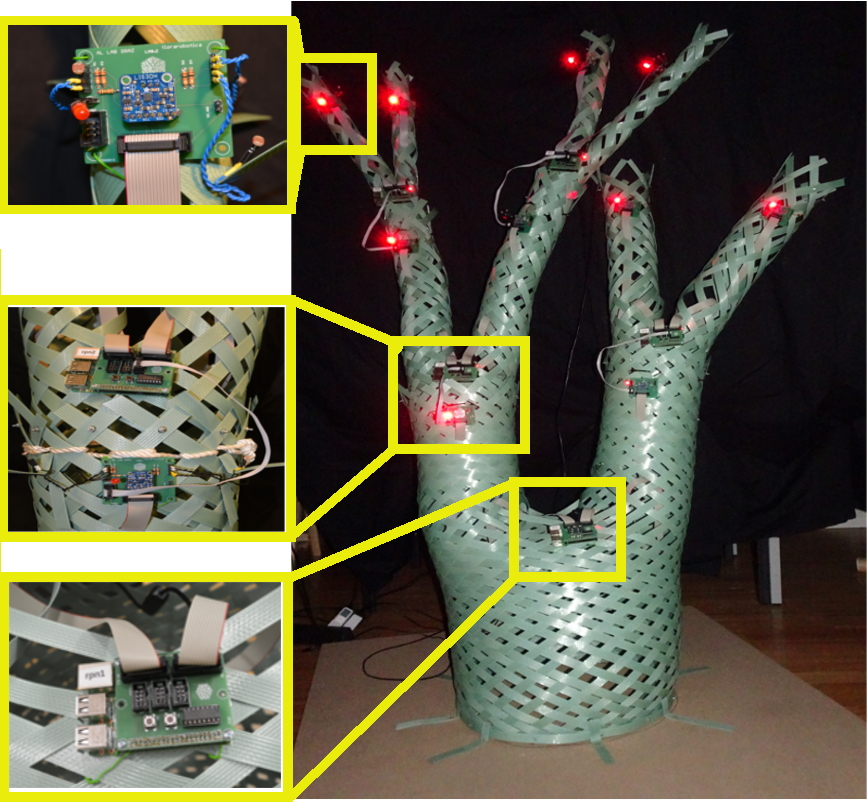}
\caption{Structure built from five Y-shaped modules. {\it Top inset}: Leaf node board attached to an arm of a L2-module. On the right side, one of the special pieces of PET straps is shown supporting one of the four photoresistors. {\it Middle inset}: Connection between two modules. The braids are tied together with a white rope to enforce continuity of the yarns. The LN of the parent module is connected to the RN of the child with a ribbon cable. {\it Bottom inset}: Root node of the \emph{RPN1} module.}
\label{fig:e-braid}
\end{figure}

\subsubsection{Root Node: RPi add-on board}


Fig.~\ref{fig:rpn-root-board} shows the assembled realization of the RN board (without RPi).
It serves as the interface between the RPi and the two LN boards, and additionally allows to connect (up to three) parent modules.
The mechanical design of the RN enables it to be stacked and mounted with screws onto an RPi\footnote{The mount holes are compatible with RPi models A+, B2, 3 and Zero.}, allowing for a solid setup.
Two 16-pin IDC connectors, namely, ``to Leaf 1'' and ``to Leaf 2'' as seen in Fig.~\ref{fig:rpn-root-board} connect the RN to its two LNs.
Each of these connectors connects four analog devices, an accelerometer, one LED and the connections to a potential child module from each of the LNs to the according pins on the RPi or, in the case of the analog devices, to pins on an {$8$-channel} ADC\footnote{MCP3008: https://cdn-shop.adafruit.com/datasheets/MCP3008.pdf}, which sits on the RN.
For convenience, the RN also contains two additional buttons.



\begin{figure}[htb]
\centering
\includegraphics[width=0.48\textwidth]{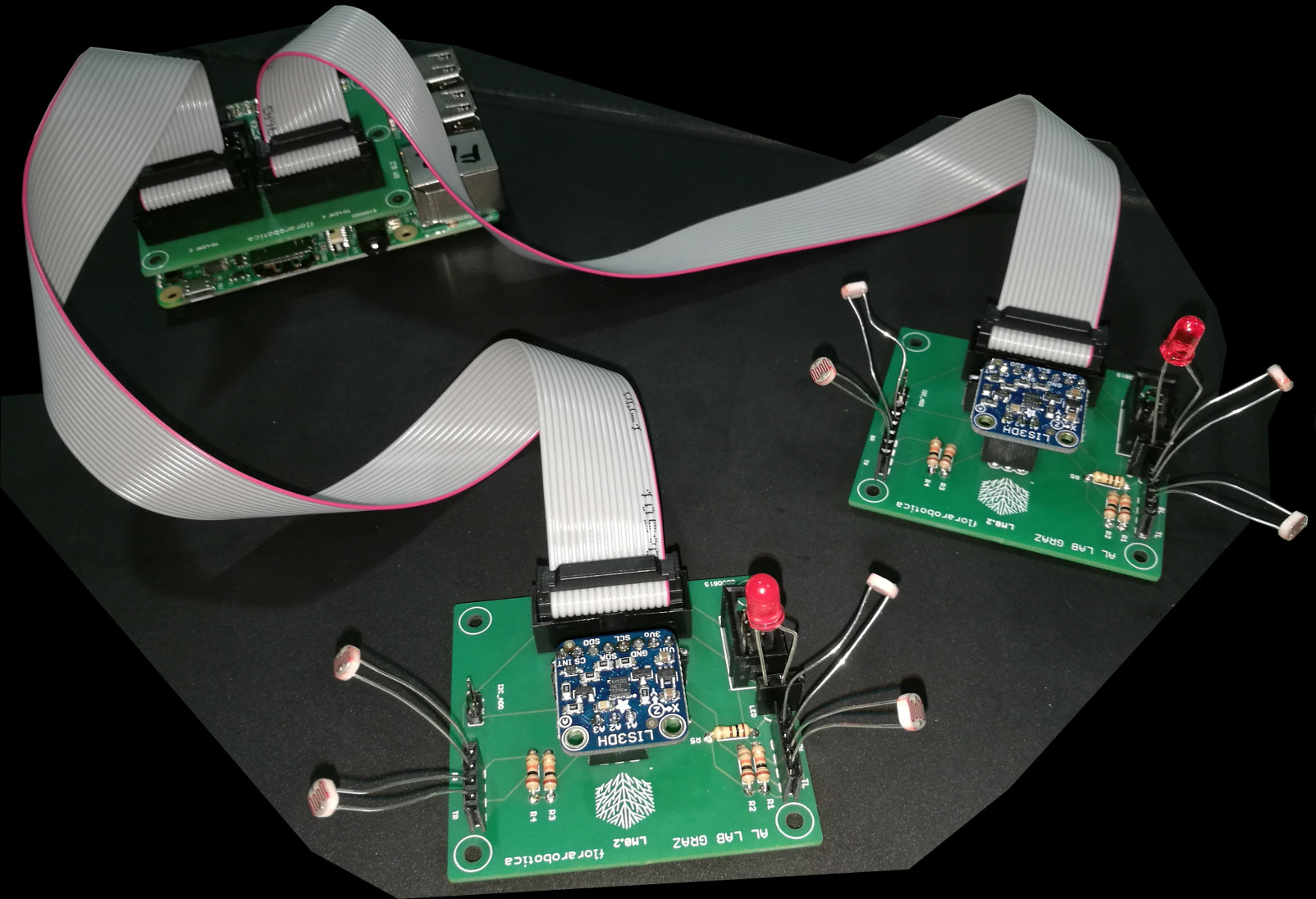}
\caption{The electronic parts of an RPN-module. In the front, the two leaf nodes with sensors are shown, in the back is the Raspberry Pi with the root node add-on board.}
\label{fig:rpn_module}
\end{figure}

\subsubsection{Leaf Nodes: connected sensor-boards}


Fig.~\ref{fig:rpn-leaf-board} shows an assembled 
Leaf Node (LN) which functions as the sensor unit at the top of each arm of an RPN module.
The LN has a 16-pin IDC connector
that allows to connect to the RN by a ribbon cable as seen in Fig.~\ref{fig:rpn_module}.
The sensor bundle on each leaf module consists of an accelerometer sensor\footnote{Adafruit LIS3DH: https://www.adafruit.com/product/2809} and four photo resistors\footnote{CdS photoresistor ``EXP-R05-735 Sparkfun LDR'': https://dlnmh9ip6v2uc.cloudfront.net/datasheets/Sensors/LightImaging/SEN-09088.pdf} for sensing ambient light.
The accelerometer is connected to the RPi through an I\textsuperscript{2}C interface.
Since there are two LNs connected to each RN, the I\textsuperscript{2}C devices are enumerated in hardware using the jumper present on the LN as seen in Fig.~\ref{fig:rpn-leaf-board}.
The photo resistors are connected through the 16-pin IDC connector to the ADC of the RN.
The 6-pin IDC connector allows to connect to a child RPN module with a small ribbon cable(see Fig.~\ref{fig:e-braid}).
A red $5$~mm LED indicates the amount of $R$ at the leaf by blinking faster for larger amounts.



\subsection{Braided modules}

Y-shaped braided structures are used as modules to represent the growth process.
The modules are produced by hand-braiding 19 mm wide and 1 mm thick PET strappings\footnote{AMSA Verpackung GmbH (Austria). The PET strappings have a 406 mm core and a traction of 800 kg.}. 
Such material specifications were selected in order to obtain sufficient structural rigidity for the modules to be able to sustain themselves while allowing the desirable deformation in the growth process. 

Each module starts from a planar base ring at its base (the \emph{root}) which hold an even set of yarns $N$, and subsequently branches into two ($N/2$) sets of yarns (the \emph{leaves}).
The number of yarns $N$ determines the width of the base ring and reflects on the width of each module, and by consequence, on the width of its respective branches.
At the leaves' ends, the yarns are bolted together pairwise, with the threaded ends of the bolt facing outside the braided tube (this is to reduce obstacles when connecting modules).
Finally, the base ring (which is helpful during hand-braiding) is removed and the yarns bolted together such that the threaded ends of the bolts are inside the tube.
Removing the ring removes a constraint on the diameter.

The electronic parts are then attached to the surface of the braided module using insulated wires.
The RPi with the root node board is placed just below the forking point of the braid, the leaf node boards are fixed near the tops of the arms at a junction just below the bolted circle.
The light sensors are placed on special pieces of PET-straps that are woven into the braid and protrude near the top.
The $16$-pin bus cables connecting root and leaf nodes can also be woven into the braid.

To connect two modules, we extend the diameter of the receiving arm of the parent and/or compress the foot of the child to smoothly insert one into the other until the root node of the child almost touches the parent's braid.
Then we tie them together with a rope, junction for junction, to achieve a continuity of the braid at that height.
The diameters of the two connected tubes are now constrained to be equal to each other, but together they can still expand and contract.

In total, five modules were constructed.
The single base module at level $0$ (L0) needs $48$ yarns, each of $160$ cm length and branches into two $24$-yarn arms.
It has a base diameter of about $45$ cm and a height of $70$-$100$ cm, depending on the forces acting on it.
We use \emph{RPN1} to denote this specific braided module including its electronic parts.
Its two leaf-nodes (at the top of the arms) are \emph{RPN1-1} and \emph{RPN1-2}.
We constructed two L1 modules (\emph{RPN2} and \emph{RPN3}) from $24$ yarns of $120$ cm length as well as two L2 modules (\emph{RPN4} and \emph{RPN5}) from $12$ yarns of $80$ cm length.
The L0 module (\emph{RPN1}) was fixed to a chipboard ($100\times100\times2.25$ cm) to ensure basic stability of the structure.
See Fig.~\ref{fig:e-braid} for a photo of all five modules forming a common structure.

\section{Software Implementation}



\subsection{Local communication}
To allow neighboring RPis to exchange values, we use GPIO pins with the
\emph{gpiozero}\footnote{https://github.com/RPi-Distro/python-gpiozero.
Version 1.4.1 or later is required, as before there was a crucial bug in gpiozero.SmoothedInputDevice, where the median of queued values was taken instead of the arithmetic mean (this makes sense for jittery sensors, but not in our case).
We have raised the issue on GitHub and it has been fixed in v1.4.1 by exposing the choice to the user.}
library for Python\footnote{https://www.python.org (version 2.7.13)}. 
Three wires connect two Raspberry Pis: one for receiving, one for transmitting and one ground.
This requires two GPIO pins per connection on each RPi.
A transmitter pin on one RPi is connected to a receiver pin on the other RPi and vice-a-versa.
This way, RPis can safely be plugged together and unplugged during runtime, with changes being noticed on all involved RPis.

The sender pin is configured as a \emph{gpiozero.PWMOutputDevice}\footnote{https://gpiozero.readthedocs.io/en/stable/\\api\_output.html\#pwmoutputdevice} that performs pulse-width modulation.
We set the initial duty-cycle to~$0.2$ (at a frequency of~$100$~Hz), such that a receiver is guaranteed to detect a connected neighbor at any time, even when the VMC-node sends the value $0.0$.
The duty cycle thus varies in the interval~$[0.2, 1.0]$, transmitting values that are scaled to the range~$[0.0, 0.8]$ and added to the basic duty cycle.

The receiving pin is set up as a \emph{gpiozero.SmoothedInputDevice}\footnote{https://gpiozero.readthedocs.io/en/stable/\\api\_input.html\#smoothedinputdevice} that is running a queue in the background that continuously polls (and stores) the pin's value.
A single value can either be $0$ or $1$, depending where in the PWM duty cycle we measure.
To deduce the sender's value reliably (accuracy:~$\pm 0.01$), we use a large queue length of~$5000$, i.e., the last $5000$ values will be averaged every time we read the receiver pin.
The receiver's threshold is set to~$0.1$, above which value the device will be considered \emph{active}, indicating that a connection to another Raspberry Pi is established and live.
Since the sender's basic PWM duty cycle is~$0.2$, we're on the safe side and reciprocal detection is guaranteed.



A major limitation of this simple (and `fuzzy') communication is that the signal needs  to be encoded within the interval $[0, 1]$.
Since $S$ from different leaves is added together on its way to the root, we have to define a maximal amount of $S$ that a leaf can produce.
This amount depends on the number of modules used for growth.

\subsection{VMC implementation}

Following an OOP-approach, for each of the five interfaces to neighbouring RPN-modules, we implement a class \emph{Neighbour} that includes all the methods and attributes necessary for local communication between two RPN-modules.
For running the LNs and managing the interface to a child module, we derive a subclass \emph{Child}, that contains the necessary interfaces to all the sensors\footnote{
For reading the LIS3DH accelerometer, we use code from Matt Dyson's GitHub-repository (https://github.com/mattdy/python-lis3dh), to which we have contributed by updating dependencies.}
attached to the leaf nodes.
The Child class also contains the methods to either generate (if the respective LN represents an unconnected leaf) or receive and transfer successin $S$ from a child module,
as well as the methods to adjust vessel thickness $V$ and for receiving resource $R$ from a parent and transmitting it over the wire.
The \emph{Parent} class, which is also derived from the Neighbour class, is simpler, as it does not contain any additional sensors.
It has a method to receive $R$ from (up to three) connected parent RPN-modules, or, if unconnected, generate it.
It also has a method for distributing the $S$ coming from the leaves or children to connected parent RPN-modules.

The parameters of the VMC and the configuration of experiments are stored in two config-files, an immutable one containing the VMC-constants (i.e., the `genome' of the VMC-system) 
and another one that stores the current state of a module and all its interfaces.
The latter one allows the software to resume from where it stopped after a system-crash, a manual restart or even the replacement of a module.

Upon inception, the main script first parses the two config files and initializes the {\it Parent}- and {\it Child}-interfaces with those values.
It then enters an infinite loop, during each iteration of which the VMC-algorithm is worked through first.
The chronology of events goes as follows:

\begin{enumerate}
    \item Receive $R$ from all connected parents or else generate it.
    \item Receive $S$ from all children (leaves or child-modules).
    \item Adjust $V$ according to the amount of $S$ received from each child.
    \item Distribute $R$ (received in step~$1$) to the children according to their relative $V$.
    \item Distribute the $S$ to the parents according to the amount of $R$ received from each (in step~$1$).
\end{enumerate}

After this is done, all the sensors are each read for themselves.
This is only done for sending them off to the visualization (see below) and for debugging reasons.
Finally, all information from a single iteration is stored as a row in a dedicated CSV-file, the mutable config-file is updated and the information is published to WiFi (also see below).


Between updating its received and altering its transmitted values, the main script waits for a random time (within bounds) before the next iteration, to guarantee asynchrony between the modules.
Because it takes a while for the large queues at the receiving pins to fill up, we also do this to minimize systematic bias between any two connected RPN modules.

\subsection{Wireless communication and visualization}

Each RPN-module continually publishes its state to the local WiFi network using ZeroMQ\footnote{http://zeromq.org (asynchronous messaging library)} \emph{publisher} sockets.
A PC in the same network runs a \emph{subscriber} socket (the publisher's counterpart) that listens for messages from all RPN modules and aggregates them into an array that is stored as a CSV file at regular intervals.

Since an RPN module is unaware of the identities of its neighbouring modules, we manually maintain a file that stores the connectivity information.
This file is read at each iteration of the WiFi-listener, and the information is appended to the according rows in the exported CSV file.

\begin{figure*}
\centering
\includegraphics[width=\textwidth]{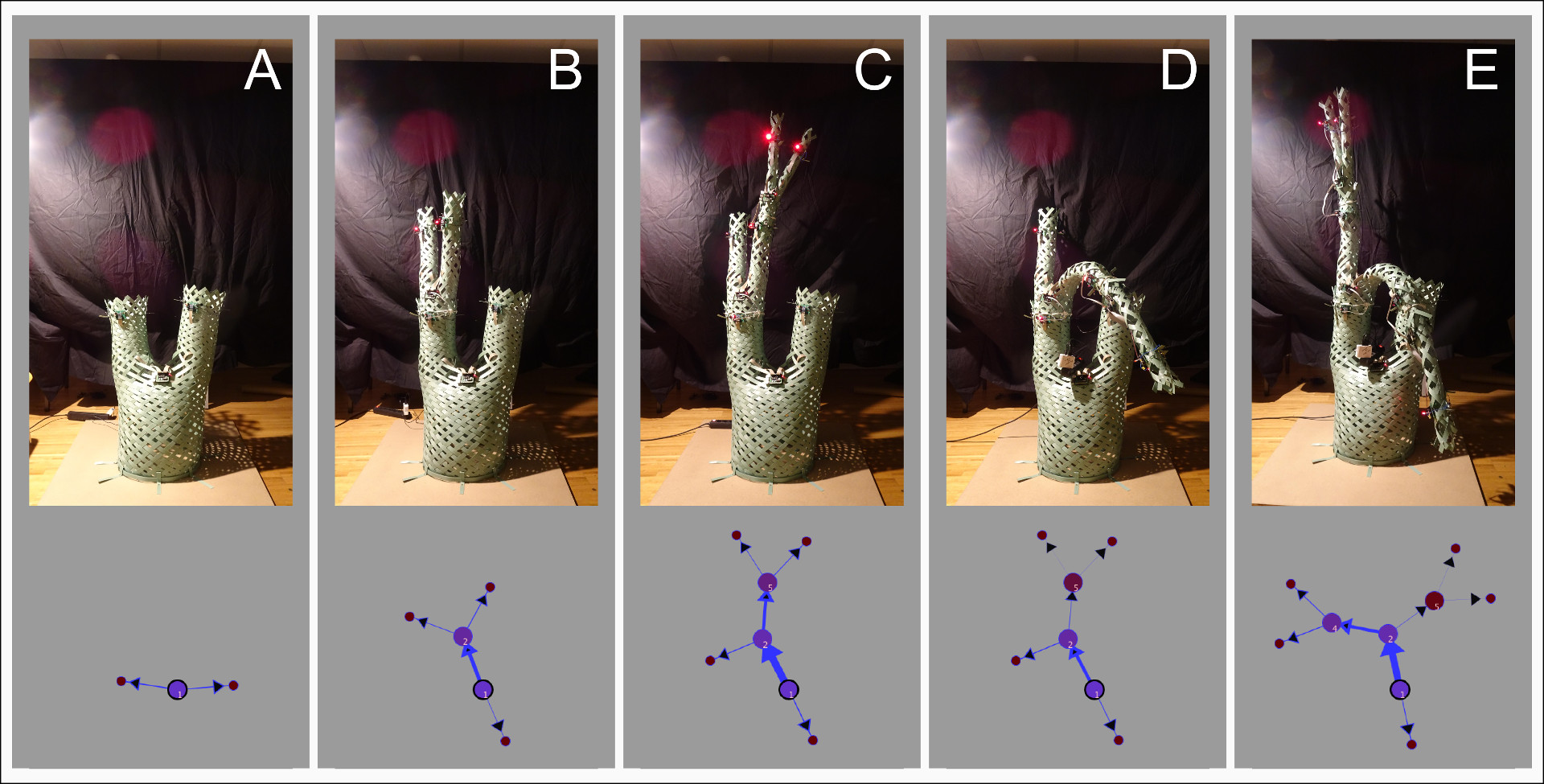}
\caption{Interactive growth guided by VMC. In the top row, photos of the different states of the real-world braidRPN-system are shown. The corresponding VMC-graphs are plotted in the bottom row. The thickness of the blue arrows between nodes indicates vessel-thickness $V$, the color of nodes represents the amount of resource, with purple being the maximal and dark-red minimal amount of resource $R$. The nodes are labelled by numbers that correspond to the modules' RPN-hostnames.}
\label{fig:exp2}
\end{figure*}

This CSV is in turn read by a program
(written in \emph{Processing}\footnote{
https://processing.org/ (version 3.3.7)
}
)
that visualizes the network and interactively outputs information on the different modules (see Fig.~\ref{fig:exp2} for the graphs).
  



\begin{figure}[htb]
\includegraphics[width=0.48\textwidth]{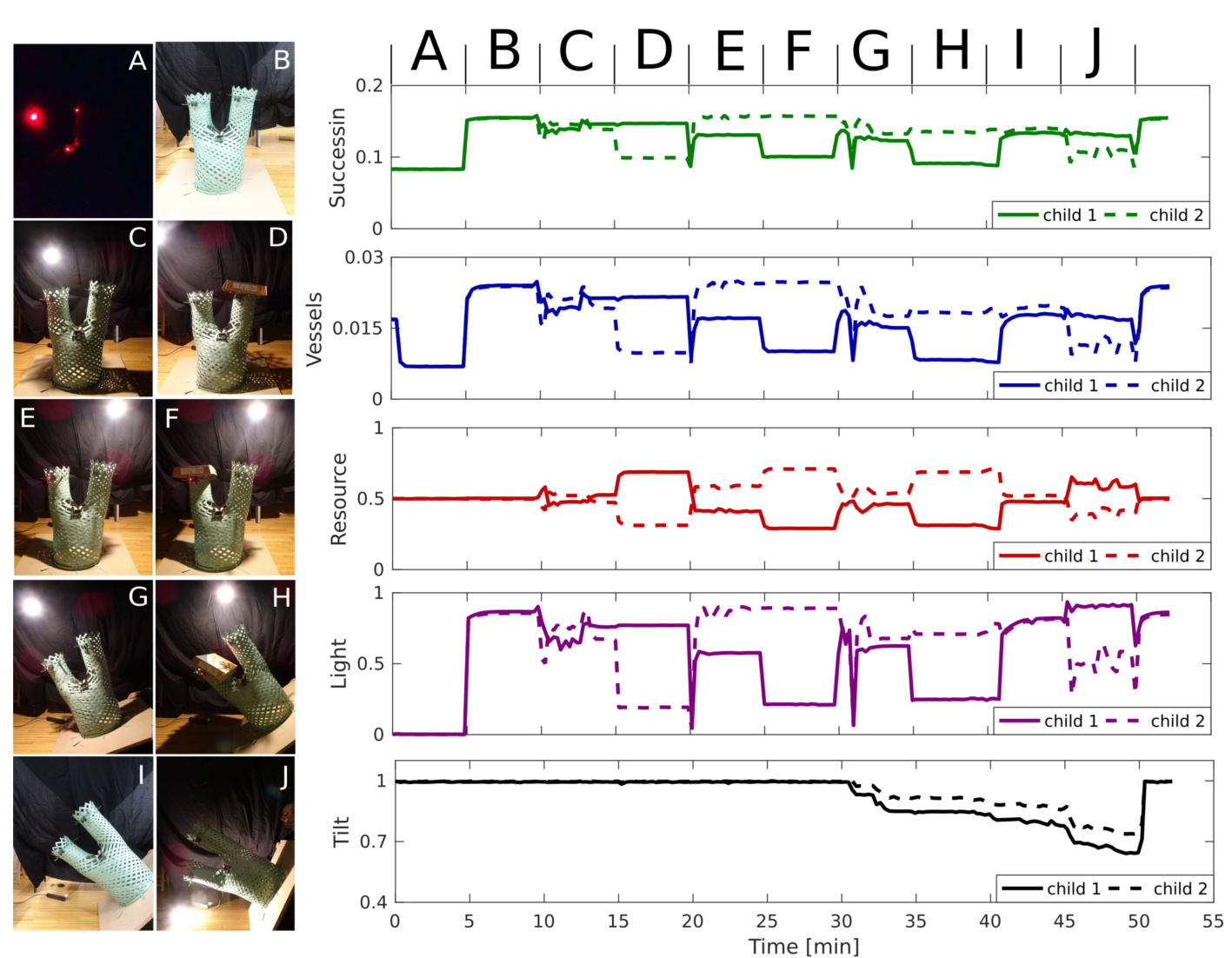}
\caption{
Characterizing a single module's behaviour.
On the left, photos of the L0 base module (\emph{RPN1}) in different states (\emph{A}-\emph{J}) are shown.
The five graphs on the right side show the output of the VMC variables ($S$, $V$ and $R$), as well as the effective sensor readings that determine those variables throughout the experiment.
The $5$ min slots corresponding to photos \emph{A}-\emph{J} are indicated above the graphs; note that the vertical axes of each graph is scaled individually.
{\it A}: darkness, {\it B}: room light, {\it C}: lamp left, {\it D}: lamp left + shade right, {\it E}: lamp right, {\it F}: lamp right + shade left, {\it G}: tilt left + lamp right, {\it H}: tilt left + lamp right + shade left, {\it I}: tilt left + room light, {\it J}: tilt left + user-targeted light. The solid line indicates the left leaf \emph{RPN1-1}, while the dashed line represents \emph{RPN1-2}.
}
\label{fig:exp1}
\end{figure}

\section{Results}

To assess the performance of the system, we first characterized the behaviour of a single module under a set of defined environmental situations.
Then we conducted an interactive growth experiment, where we followed the decisions of the VMC-algorithm on where to attach new modules, but critically impacted the system
to show the redistribution of resources within the network to adapt to a global challenge.



\begin{table}[tbp]
\centering
\caption{Parameterization of the VMC used in this work. For both the $\omega$- and $\rho$-parameters, subscripts $c$, $\phi$ and $\lambda$ denote the constant, tilt-angle- and light-associated weights respectively (see equations in Fig.~\ref{fig:VMC}).}
\label{tab:vmc-constants}
\begin{tabular}{|llllllll|}
\hline
$\omega_{c}$           & $\omega_{\phi}$         & $\omega_{\lambda}$      & $\rho_{c}$              & $\rho_{\phi}$           & $\rho_{\lambda}$      & $\alpha$                & $\beta$                \\ \hline
\multicolumn{1}{|c}{0} & \multicolumn{1}{c}{0.5} & \multicolumn{1}{c}{0.5} & \multicolumn{1}{c}{0.9} & \multicolumn{1}{c}{0.1} & \multicolumn{1}{c}{0} & \multicolumn{1}{c}{0.9} & \multicolumn{1}{c|}{2} \\ \hline
\end{tabular}
\end{table}

Throughout the experiments, we used the parameter set presented in table~\ref{tab:vmc-constants}.

\subsection{Characterization of a single RPN-module}

The large base module \emph{RPN1} was used to record the effects of light and gravity on the balance of resources between its two branches (see Fig.~\ref{fig:exp1}).
In general, the behaviour was as expected.
Production of the hormone $S$ varies with the conditions at the leaves, with more light and a more upright posture leading to more $S$.
The graph of vessels $V$ closely resembles that of $S$, yet differences in $S$ are amplified in $V$ according to the equation \emph{Update of a Vessel} in Fig.~\ref{fig:VMC}.
Since the resource $R$ is distributed according to the \emph{relative} amount of $V$, the value of $V$ only matters in regard to its sibling branches.
That is, if both branches have similar $V$, each receives about half of the $R$, and only when they differ in $V$ does the distribution get skewed.
In Fig.~\ref{fig:exp1}, we see that during the symmetric states \emph{A} (darkness) and \emph{B} (room light), $S$ and $V$ dramatically change, but that there is no impact on the distribution of $R$.
We then employ a mobile lamp and a shade (cardboard), to tip the balance both ways (states \emph{C}-\emph{F}).
First, we favour the leaf \emph{RPN1-1}, and then \emph{RPN1-2}, the results of which can be read from the graph of $R$ in Fig.~\ref{fig:exp1}.
Since the lamp's beam is not very focused, placing the shade has a stronger impact than moving the lamp.
In states \emph{G}-\emph{J} we slowly tilted the board ever more leftward
(in the bottom graph of Fig.~\ref{fig:exp1}, a value of $0.5$ would correspond to a tilt of $90$\degree).
Tilting the board to the left naturally affects the left leaf slightly more.
In the final state~\emph{J}, we counter these effects by manually lighting the sensors of~\emph{RPN1-1}.


\subsection{Interactive growth guided by VMC}


Fig.~\ref{fig:exp2} shows the course of an experiment in which the VMC-algorithm indicated where to add the next module.
The mobile lamp was placed on the left side and wasn't moved in this experiment.
Initially, the two leaves of the L0 base module (\emph{RPN1})
produce similar amounts of successin ($S$), with the left one (\emph{RPN1-1}) being slightly favoured ($54\%$ vs.~$46\%$)
(Fig.~\ref{fig:exp2}A).

Following the algorithm, we connect an L1-module (\emph{RPN2}) to the favoured leaf \emph{RPN1-1} (Fig.~\ref{fig:exp2}B), turning the latter into an intermediary node.
Next to being shaded by \emph{RPN2}, the now single leaf of \emph{RPN1} is also outperformed by the combined $S$-production of the two new L1-leaves.
With its thus larger vessels, the branch toward \emph{RPN2} attracts $\approx86\%$ of resource ($R$), depleting the single L0-leaf.
This behaviour is analogous to \emph{apical dominance} in plants, where a successfully growing tip suppresses the outgrowth of branches behind it~\citep{muller2015cytokinin}.

The three existing leaves \emph{RPN1-2}, \emph{RPN2-1} and \emph{RPN2-2} respectively receive $14\%$, $39\%$ and $47\%$ of $R$, so we add the L2-module \emph{RPN5}, to the leaf~\emph{RPN2-2} (Fig.~\ref{fig:exp2}C).
This further enforces the vessels between \emph{RPN1} and~\emph{RPN2} as well as depletes the resources of the left-behind leaf~\emph{RPN2-1}.

It is interesting to note that at this point, the single leaves on the lower levels (\emph{RPN1-2} and \emph{RPN2-1}) both produce more $S$ (and thus have more vessels) than either of the two new top level leaves (\emph{RPN5-1} and \emph{RPN5-2}), yet the latter both receive more $R$ than either of the lower ones.
The top leaves produce less $S$ because they are already above the placed mobile lamp and therefore receive less light than the lower ones.
But, again, their combined $S$ is sufficiently larger than the $S$ from the leaf below, allowing them to channel more $R$ toward both of them.
This dominance propagates down the stem as all well-performing single leaves' $S$ is added to vascularize the stem such that it is capable of supplying the demand of the ever more distant growing apex.

In the state shown in Fig.~\ref{fig:exp2}C, leaf \emph{RPN5-1} commands most $R$~($37\%$), so that's where we would add the next module according to the VMC-algorithm.
Instead, we tilt the \emph{RPN5}-branch such that gravity makes it bend over (Fig.~\ref{fig:exp2}D).
With the tips facing downward, $S$-production is roughly halved, leading to an almost ten-fold reduction in vessels $V$(according to the parameter set (table~\ref{tab:vmc-constants}) and the equation \emph{Update of a Vessel} in Fig.~\ref{fig:VMC}).
Now leaf \emph{RPN2-1} harnesses $51\%$ of $R$ (up from $21\%$) compared to only $30\%$ for the whole of \emph{RPN5}.
Being upright and well-supplied by light, \emph{RPN2-1} is now a far more attractive option to continue the growth of the structure.

Finally, we add the L2-module \emph{RPN4} to leaf \emph{RPN2-1} (Fig.~\ref{fig:exp2}E) to demonstrate the take-over of apical dominance: \emph{RPN4}
now commands $\approx75\%$ of the total available $R$.

\section{Discussion and Conclusion}
Plant growth models have been explored extensively in-silico.
A key benefit of in-silico modelling is the ability to model continuous incremental accumulation of material \textendash{} a process which is challenging for investigation through physically embodied artificial systems.
This paper has introduced a novel approach to plant-inspired embodied artificial growth that combines the benefits of material continuity through braiding, with a distributed and decentralised Vascular Morphogenesis Controller.
We have demonstrated how this artificial system can determine adaptations required to better situate itself to exploit environmental resource, and to signal necessary morphological changes to a human agent.
A clear limit of this work is that artificial growth by material accumulation does not occur autonomously.
However, in the context of design fields, such as Architecture, the incorporation of human agency in the steering of self-organised systems can offer a number of advantages over fully autonomous systems.
Maintaining the `human in the loop' establishes a rich design space supporting the negotiation between pre-determined (but, being anticipatory, always under-informed) objectives and self-organised embodied action that can continually suggest modification against `real-world' conditions.
In this context, our approach offers a clear contribution by providing an integrated material and computational basis to support continuous design coupled directly to embodied spatial adaptation\footnote{The authors are happy to provide all the recorded and aggregated data, the code producing it, the PCB-design files, as well as (timelapse-)video recordings of the experiments.}.

\section{Acknowledgements}


This work was supported by EU-H2020 project `{\it flora robotica}', no.~640959.
We further thank Stefan Sch{\"o}nwetter-Schistek-Fuchs, Sanel Durakovic, Asya Ilg{\"u}n, Daniel E. Moser, Johannes Rabensteiner, Martin Stefanec and Nikolaus Sabathiel for valuable ideas in conceiving and practical assistance with building and documenting the hardware.

\footnotesize
\bibliographystyle{apalike}
\bibliography{vmc} 

\end{document}